# $M^2$Deep-ID: A Novel Model for Multi-View Face Identification Using Convolutional Deep Neural Networks

Sara Shahsavarani[a], Morteza Analoui[a] ∗, and Reza Shoja Ghiass[b] ∗∗

[a] *School of Computer Engineering, Iran University of Science and Technology, Tehran, Iran*
[b] *Université Laval, Quebec City, Quebec, Canada*

## ABSTRACT

Despite significant advances in Deep Face Recognition (DFR) systems, introducing new DFRs under specific constraints such as varying pose still remains a big challenge. Most particularly, due to the 3D nature of a human head, facial appearance of the same subject introduces a high intra-class variability when projected to the camera image plane. In this paper, we propose a new multi-view Deep Face Recognition (MVDFR) system to address the mentioned challenge. In this context, multiple 2D images of each subject under different views are fed into the proposed deep neural network with a unique design to re-express the facial features in a single and more compact face descriptor, which in turn, produces a more informative and abstract way for face identification using convolutional neural networks. To extend the functionality of our proposed system to multi-view facial images, the golden standard Deep-ID model is modified in our proposed model. The experimental results indicate that our proposed method yields a 99.8% accuracy, while the state-of-the-art method achieves a 97% accuracy. We also gathered the Iran University of Science and Technology (IUST) face database with 6552 images of 504 subjects to accomplish our experiments.

.

## 1. Introduction

### 1.1. background

Face recognition is a well-studied topic in the literature. However, its success is limited due to several factors such as illumination, facial pose, and facial expression. Each of this constraints has opened a new research direction in the literature. As an example, infrared face recognition was proposed to cope with the illumination dependency problem [1]-[4], while 3D face recognition was suggested to tackle both facial pose and expression [5]-[7].

Face recognition using 2D or 3D imaging systems is a long-term problem in the computer vision corpus. The 2D face images of an individual may be captured from the different view angles. Therefore, they may cause different appearances of a face for the same individual. On the other hand, identifying faces in different orientations is one of the advantages of 3D face recognition systems. However, this kind of face recognition systems faces a few issues. Lack of the three-dimensional face databases, long acquisitive time of creating a three-dimensional model and cost of three-dimensional scanners are some examples of those issues that persuade the researchers to employ two-dimensional facial images for identification. Moreover, the conventional face recognition methods required a powerful feature extraction technique to extract critical and effective features. Furthermore, these methods usually used linear classifiers [8]. So they did not have reliable results if the problem would be complex. Since deep learning emerged, due to its ability in automatic feature extraction and non-linear classifying, face recognition methods have been improved significantly [9]-[13]. As an example, the DeepID [12] model has become the golden standard for DFR using 2D images. Despite these improvements, there are few methods for multi-view face recognition.

In [14]-[16] multiple networks are used to deal with the image faces in different orientations. For instance, in [15] the authors modified the pose to the frontal (0°), half-profile (40°) and full-profile views (75°) and then addressed pose variation by multiple pose networks. In [14] a multi-view deep network (MvDN) consisting of view-specific sub-networks as well as common sub-networks was designed. In MvDN, the first sub-network removes view-specific variations, and the second one obtains common

---

representations. Wang et al. [16] used coupled SAE for cross-view face recognition and employed a deep neural network to model the view distribution. The authors reported quite promising performance.

*1.2. Motivation*

To design our multi-view representation method, we were inspired by several works in the literature. As an example, we were motivated by the concept of Structure from Motion (SfM) [17], [18]. SfM is a technique to recover the three-dimensional structure from two-dimensional image sequences. In this context, we focused on the idea of feeding multiple 2D views of the subjects into the neural network and let the network calculate a more abstract representation of the features extracted from the 2D views.

The DeepID model (See Fig. 2) is the golden standard in the domain of DFR. However, we noticed that this model is designed to identify frontal faces. This inspired us to modify this model and integrate it in a more general framework. Thanks to our specific design, the modified model should be able to classify multi-view images.

*1.3. Contributions of the proposed method*

In this paper, we present a novel approach for multi-view face identification using deep learning. As a face has a three-dimensional nature, not every facial feature can be sensed due to self-occlusion when projected to the camera's image plane. On the other hand, it seems reasonable to obtain as much information as possible for each subject [15], [9]. For this purpose, unlike the DeepID model, we do not rely on frontal images of the subjects only. To this end, we present a unified deep network for face recognition based on two-dimensional view-based face images. Our main contributions are the followings:

i) Two new architectures for learning multi-level and multi-view face representations are proposed.
ii) The golden standard DeepID model, which only supports frontal faces, is modified and integrated into our proposed system to identify multi-view faces too.
iii) A new method to aggregate multi-view facial images is proposed.
iv) A relatively large facial database has been acquired.

Our work is the first one that uses 2D multi-view images to achieve a single and compact representation of a 3D face using deep learning.

## 2. Method Details

In this section, we explain our methodology in details. First, a general model is presented as the baseline. Then, we explain an extension of the proposed model to extend its functionality.

### *2.1. General Model*

Fig. 3 (a) shows the general model framework. As our work extends the functionality of the DeepID model to support multi-view faces, we call the proposed model the Multi-View DeepID or MV-DeepID model. As it is demonstrated in Fig. 3 (a), the proposed MV-DeepID model contains four main modules. In the first module, a representation learning method called the multi-view representation is employed. In the second module, we design an aggregating layer. This layer aggregates the information coming from different views and produces a 2D compact representations of all views. The third module learns the aggregated features while the last module is a softmax layer for identifying each class. Each module is explained in detail next.

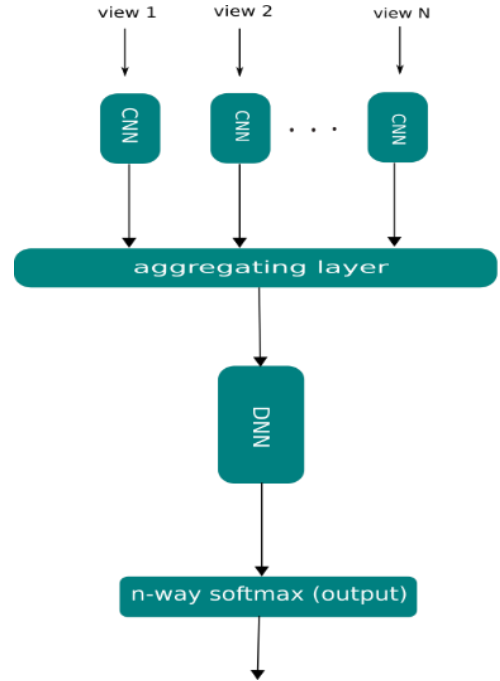

**Fig. 1**: The architecture of the proposed MV-DeepID system

#### *2.1.1. Multi-View Representation*

The first module of the proposed DFR is a face descriptor using a deep neural network with 2D view-based images as the input data. This module is basically an ensemble of some sub-networks each specialized in a specific view range.

As the DeepID model has shown very promising results for 2D face recognition, it seems reasonable to somehow integrate parts of this model in our general model. In this context, we employ the DeepID model as a pre-trained convolutional network in every sub-network (Fig. 1). While this idea shares some similarities with the well-known transfer learning technique, it has some major differences which are explained next. In this way, we borrow most of the important features learned by the DeepID model, except the last two layers. The details are shown in figure 2. The size and number of filters are like [12]. As demonstrated in this image, the input image size is 55×47×3 and the output of convolutional layers are (height, width, depth). Table 1 indicates the dimensions of the layer 1 to the layer 4.

#### *2.1.2. The Aggregating Layer*

The output layer of each sub-network in the previous module concatenate and create the aggregating layer. In other words, the aggregating layer is a shared representation of all views. So, this layer conveys more informative representations of a face employing 2D multi-view face images than each sub-network in single.

Given *N* views, the dimension of the aggregating layer, d, is:

$$d = (N \times H) \times W \times D \quad (1)$$

Where H, W, and D stand for height, width and depth of the last layer of each sub-network in the previous module.

**Table 1**: The dimensions of the layer 1 to layer 4 in MV-DeepID model

| Layer | Dimension (height, width, depth) |
|---|---|
| Layer 1 | (26, 22, 20) |
| Layer 2 | (12, 10, 40) |
| Layer 3 | (5, 4, 60) |
| Layer 4 | (4, 3, 80) |

#### 2.1.3. Learning Aggregating Layers

In this section, the representations yielding in the previous section are fed into fully connected layers. The fully connected layer is a hidden layer containing 160 neurons. The number of fully connected layers and the number of neurons in this section are set experimentally.

#### 2.1.4. Softmax Layer

We employ a 504-way softmax layer to recognize the faces. The 504-way softmax layer predicts the probability distribution over 504 different identities. The proposed multi-view convolutional neural network is learned by minimizing negative log-likelihood. Stochastic gradient descent [19] is used to optimize the cost function. In our deep architecture, ReLU activation functions [19] are used in both convolutional and fully connected layers.

### 2.2. Extension of Multi-Level MV-DeepID

In the general model, the aggregating layer was constructed using the output of each sub-network. Although, this idea was promising, it turns out that the last output layer ends up with too few neurons after several consecutive max-pooling and convolutional layers. Consequently, information may not propagate effectively. To address this problem, multi-view representations are fed to the fully connected layers in the literature [ 11, 21]. Inspired by the literature, we also use the multi-level representations in the aggregation layer and call the new model the Multi-Level and Multi-View DeepID model (or $M^2 DeepID$).

In this context, the last two outputs of the convolutional layers for each specific view range aggregates first. Next, they are fed into a fully connected layer. In this step, a large vector containing features of a 3D face of the subject followed by a fully connected layer is generated. Fig. 3(b) shows the details.

## 3. Experimental Results and Discussion

We assessed the performance of the two proposed methods on *Iran University of Science and Technology (IUST)* face database. The IUST face database contains 6552 face images of 504 people. For each individual, there are face images in 5 poses (orientations): 2 images for left view, 5 images for the center view, 2 images for the right view, 2 images for the up view, 2 images for the down view. This database was gathered in neutral illumination. Moreover, some subjects were asked to wear glasses. Images of IUST face database are in RGB format.

In training both models, some data augmentation techniques such as mirroring the images or affine transformation is used to increase the size of the database synthetically. In this database, 5 poses (orientations) per individual is used: 5 images for the left view, 5 images for the center view, 5 images for the right view, 5 images for the up view, and 5 images for the down view. The resolution of the images is 55×47 pixels.

We used a NVIDIA GTX 960 GPU for performing the optimizations. Theano [21, 22] was used for implementing the model and performing the experiments. The learning rate is set to 0.0001, while the number of neurons in the hidden layer is set to 160. The number of outputs for the softmax layer is the same as the number of the classes, i.e., 504. We used stochastic gradient descent for optimization. The activation function of all layers, except the softmax layer, is set to ReLU activation function.

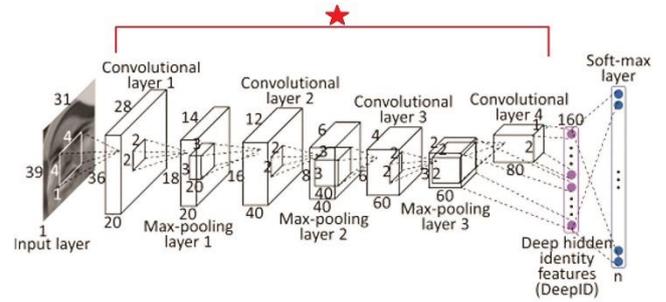

**Fig. 2**: A detailed representation of the golden Deepid model[12]. In the proposed method, we borrowed the elements, the size and the number of filters, of the starred section of the DeepID model. In the proposed method, we change the size of the input layer to 55×47×3 to be able to do further functions. So, the dimension of the convolutional layers will be changed, accordingly [12].

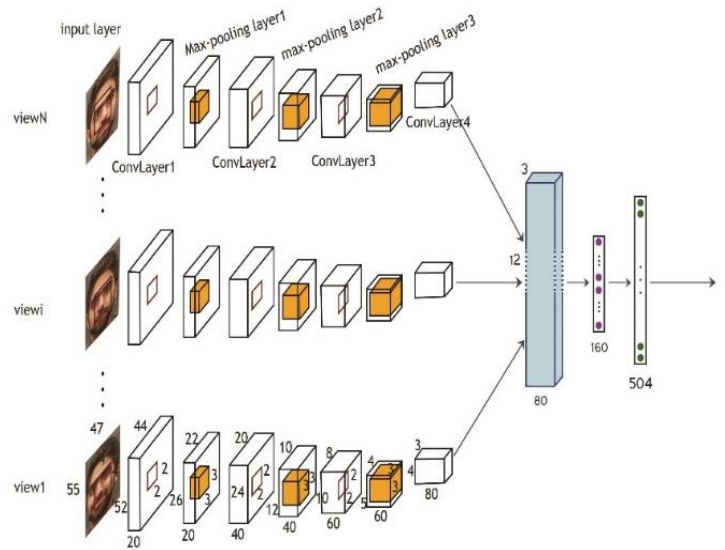

(a)

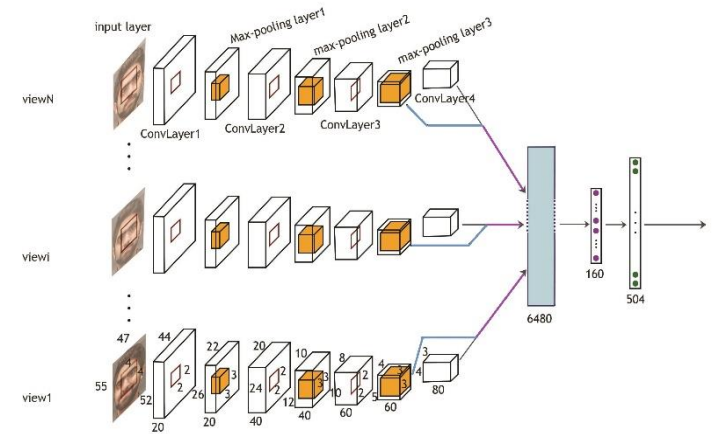

(b)

**Fig. 3**: A detailed representation of (a) the MV-DeepID, and (b) the $M^2$Deep-ID models.

In our experiments, we noticed that the aggregation layer is not capable to learn all the 5 views together (See Fig. 4 (a)). One possible reason is that we borrow the high-level feature extractors from the Deep-ID model, these filters are learned from 2D views only. Consequently, the network, as a nonlinear function, is not flexible enough to learn and classify features of the subject extracted from two completely different views (left and up for example) in the same class.

In this context, we divided the database into two sub-datasets which share more similarities. The first subset containing the left view (L), the center view (C), and the right view (R) is called the LCR sub-dataset. Similarly, the other sub-dataset contains the up view (U), the center view (C), and the down view (D). This part is called UCD. As another limit of the algorithm, we noticed that it is not possible to expect a good performance for the network in the case of profile faces or those with the extreme pose.

Fig.4 (a) shows MvDeep-ID performance using all the five views (Left, Center, Right, Up, Down). This plot indicates the error of each epoch for the training set, the validation set, and the test set. In our experiments, the error of the training set converges to 0, while the error of the validation and test sets converge to 0.2 and 0.4 respectively. On the other hand, Figs. 4(b) and 4(c) show the performance of the MV-DeepID model tested on the LCR and UCD sub-datasets respectively. The experiments show that after the 20th epoch the error of the validation set converges to 0.04, and the error of the test set converges to 0.02.

Fig.5 compares the performance of the proposed methods with the DeepID model on the training, the validation, and the test set. As we can see, the proposed methods outperform the DeepID method for multi-view faces. In other words, our model extends the functionality of the DeepID model to operate under varying poses.

Table. 2 and 3 summarizes the accuracy of each model tested on the IUST face database.

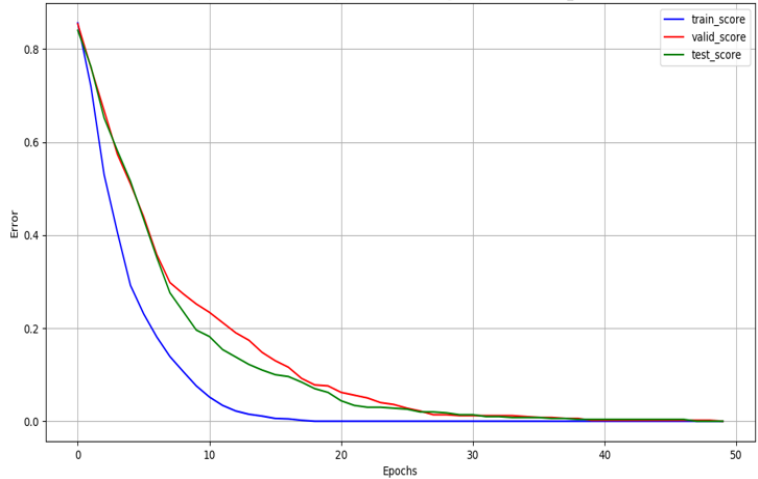

(a)

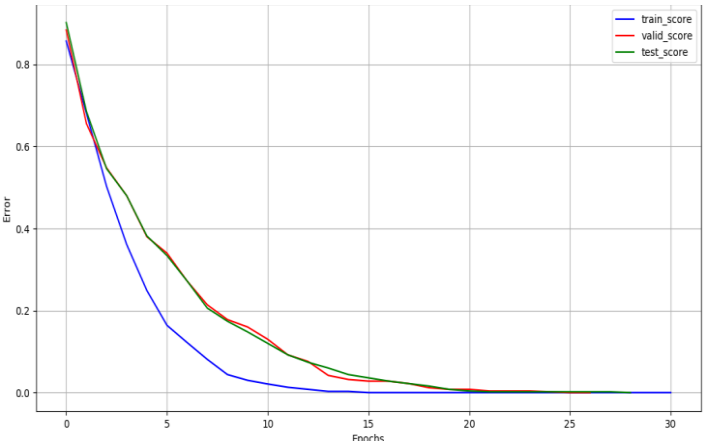

(b)

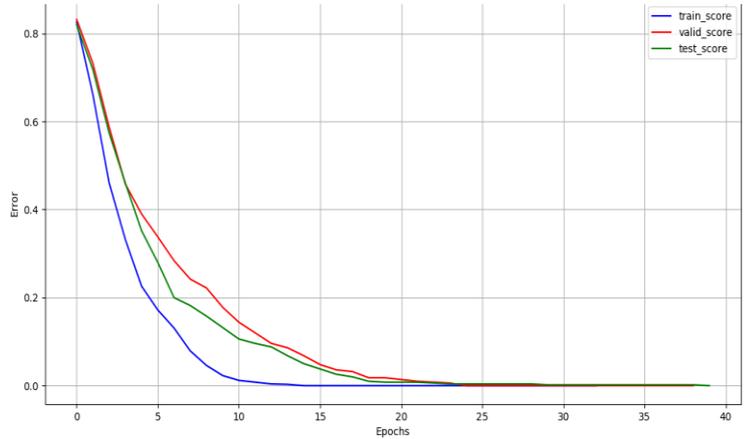

(c)

**Fig. 4**: Effectiveness of grouping the dataset into two sub-datasets: (a) The model fails to learn after 20 epochs when using all the five views, (b) The LCR sub-database, and (c) the UCD sub-dataset.

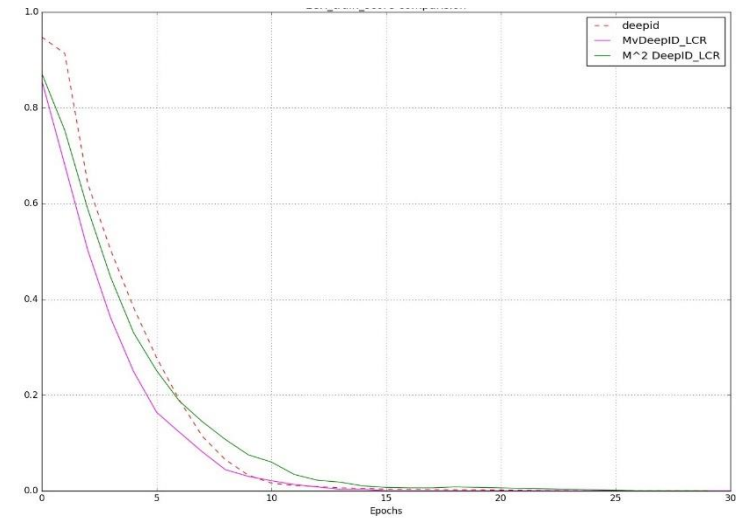

**a1**

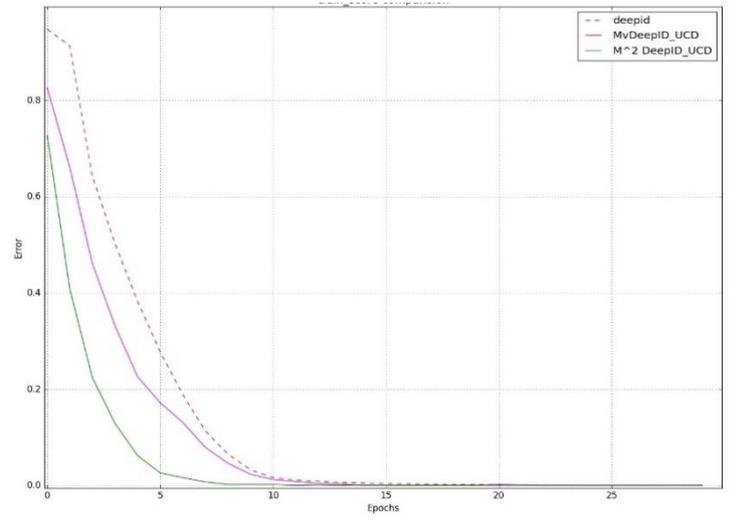

**b1**

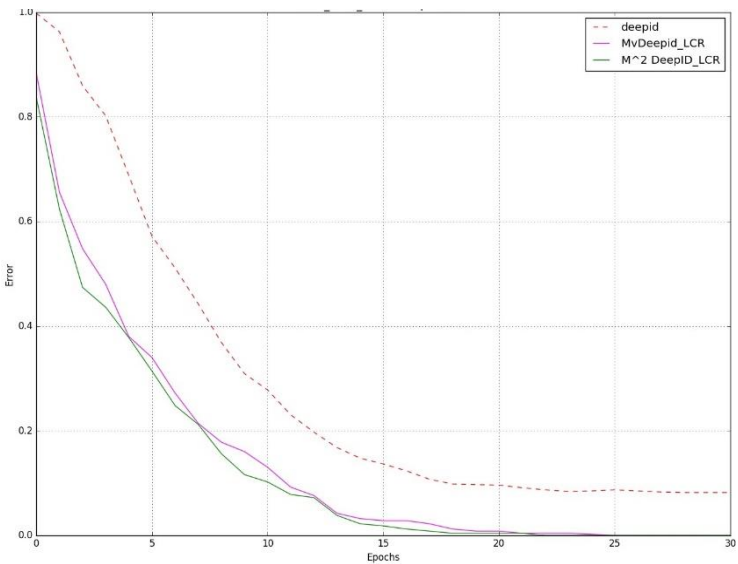

**a2**

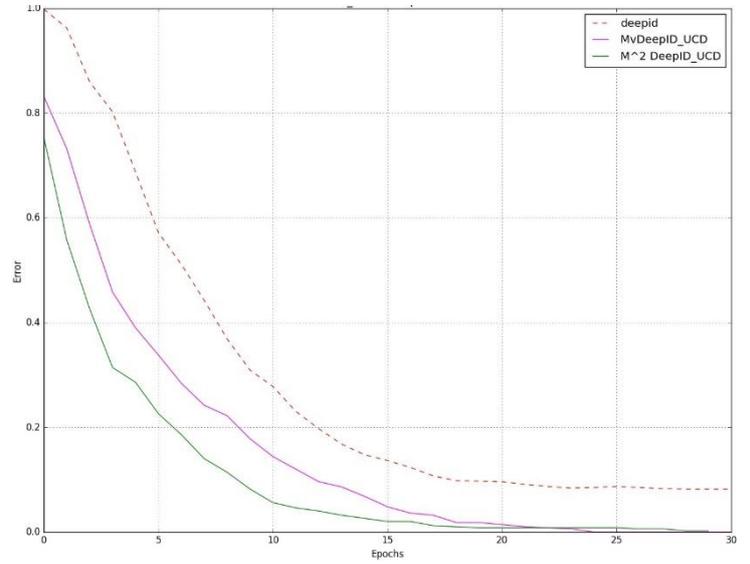

**b2**

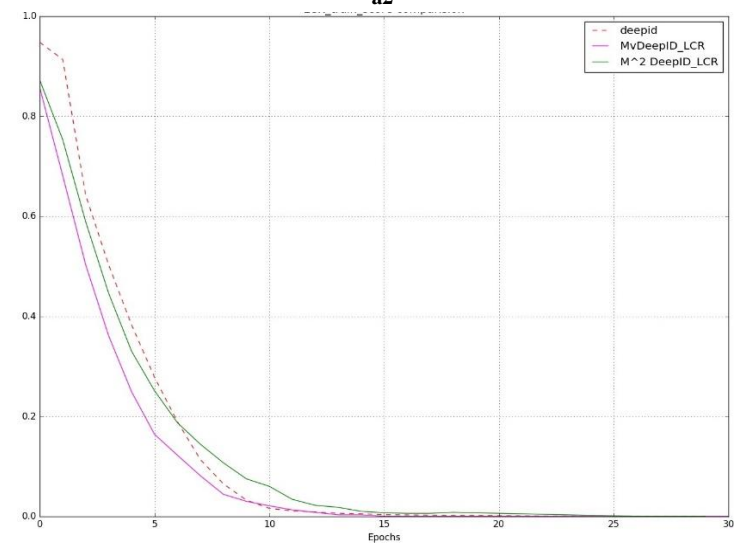

**a3**

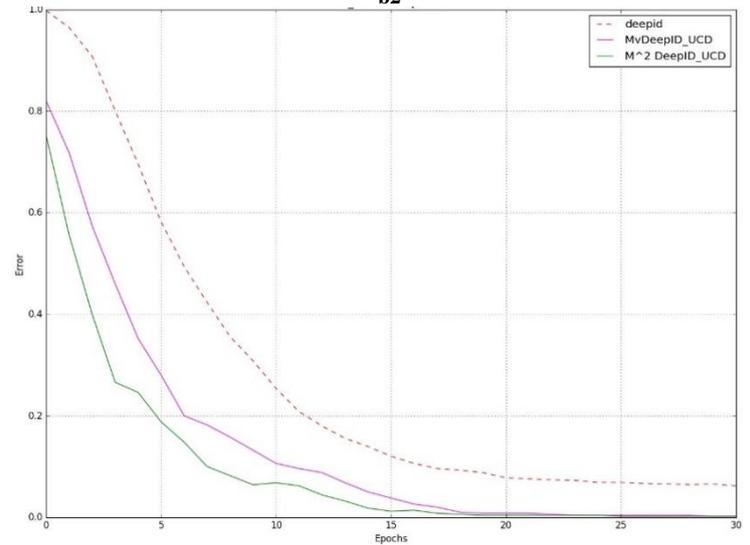

**b3**

**Fig. 5**: The performance of comparison between our proposed method and the Deep-ID: a1, a2, and a3 relate to LCR sub-database and b1, b2, and b3 relate to UCD sub-database. a1 and b1 illustrate the training set performance, a2 and b2 shows the validation set performance, and a3 and b3 indicate the test set.

**Table 2**: Comparison of the $M^2$Deep-ID and Deep-ID methods for the LCR sub-dataset

| Method | Accuracy |
| --- | --- |
| DeepID | 97% |
| MV-DeepID | 98% |
| $M^2$Deep-ID | 99.2% |

**Table 3:** Comparison of the $M^2$Deep-ID and DeepID methods for the UCD sub-dataset

| Method | Accuracy |
| --- | --- |
| DeepID | 97 % |
| MV-DeepID | 98% |
| $M^2$Deep-ID | 99.8% |

The proposed models are also compared with together and the Deep-ID model using the t-test, which is a statistical test [23].

In table 4, the t-test values illustrate how much the proposed frameworks work well in comparison to the DeepID model for both LCR and UCD sub-databases. In table 5, $M^2$ Deep-ID is compared with the MvDeepID model for both UCD and LCR sub-databases.

**Table 4:** Comparison of the DeepID model with our proposed methods through t-test values for LCR and UCD sub-databases.

| Method | t-value (train) | t-value(valid) | t-value (test) |
| --- | --- | --- | --- |
| MvDeepID-LCR | 4.8 | 9.03 | 8.29 |
| $M^2$DeepID -LCR | 3.93 | 9.55 | 7.69 |
| MvDeepID-UCD | 1.14 | 8.48 | 8.61 |
| $M^2$DeepID -UCD | 4.07 | 9.39 | 8.14 |

**Table 5:** comparison of the Mv-DeepID model and $M^2$ Deep-ID through t-test values for LCR and UCD sub-databases.

| Method | t-value (train) | t-value (valid) | t-value(test) |
| --- | --- | --- | --- |
| MvDeepID - LCR/ $M^2$DeepID -LCR | 3.65 | 4.72 | 2.90 |
| MvDeepID - UCD/ $M^2$ Deepid -UCD | 3.94 | 10.14 | 6.10 |

## 4. Conclusions

In this paper, we proposed a new multi-view Deep Face Recognition (MVDFR) system. Multiple 2D images of each subject under different views were fed into the proposed deep neural network with a unique design to re-express the facial features in a single and more compact face descriptor, which in turn, produced a more informative and abstract way for face identification using convolutional neural networks. To extend the functionality of our proposed system to multi-view facial images, the golden standard DeepID model was modified in our proposed model. The experimental results indicate that our proposed method yields a 99.8% accuracy, while the state-of-the-art method achieves a 97% accuracy when testing multi-view facial images. We also gathered the Iran University of Science and Technology (IUST) face database with 6552 images of 504 subjects to accomplish our experiments.

**Acknowledgment**

We acknowledge the members of the IUST Biometric Laboratory for supporting this project.